\documentclass[
twocolumn,
]{ceurart}

\usepackage[a4paper]{geometry} 
\usepackage{xurl} 
\usepackage[italian,english]{babel}
\usepackage{latexsym} 
\usepackage{xcolor}
\usepackage{natbib}

\begin{document}

 \copyrightyear{2023}
\copyrightclause{Copyright for this paper by its authors.
  Use permitted under Creative Commons License Attribution 4.0
  International (CC BY 4.0).}
\conference{EVALITA 2023: 8\textsuperscript{th} Evaluation Campaign of Natural Language Processing and Speech Tools for Italian, Sep 7 -- 8, Parma, IT}


\title{ACTI at EVALITA 2023:Automatic Conspiracy Theory Identification Task Overview}
\author[1]{Giuseppe Russo}[%
email=russog@ethz.ch,
]
\address[1]{ETH Z\"urich,
  Rämistrasse 101, 8092 Z\"urich, Switzerland}

\author[1]{Niklas Stoehr}[%
email=niklas.stoehr@inf.ethz.ch
]

\author[2]{Manoel Horta Ribeiro}[%
email=manoel.hortaribeiro@epfl.ch]
\address[2]{EPFL, Rte Cantonale, 1015 Lausanne, Switzerland}

\begin{abstract}
\textbf{English.} Automatic Conspiracy Theory Identification (ACTI) is a new shared task proposed for the first time at the EVALITA 2023 evaluation campaign. ACTI is based on a new, manually labeled dataset of comments scraped from conspiratorial Telegram channels and consists of two subtasks:
(1) identifying conspiratorial content (conspiratorial content classification); and 
(2) classifying content into specific conspiracy theories (conspiratorial category classification). 
A total of 15 teams participated in the task with 81 submissions. 
In this task summary, we discuss the data and task, and outline the best-performing approaches that are largely based on large language models. We conclude with a brief discussion of the application of large language models to counter the spread of misinformation on online platforms.
\end{abstract}


\maketitle

\section{Introduction}

From ancient tales of secret societies, \cite{roisman2006rhetoric} to speculation on whether the moon landing happened~\cite{moon},  belief in conspiracy theories has been prevalent throughout human history~\cite{van2017conspiracy} and has inflicted harm upon individuals and groups falsely accused of wrongdoing~\cite{douglas2019understanding}. 
For example, in the middle ages, the Blood Libel conspiracy theory falsely accused Jews of murdering Christian boys, fostering their persecution~\cite{rose2015murder}. 

Fast-forward to the digital age, the Internet has emerged as the prominent medium through which individuals are exposed to conspiracy theories \cite{sunstein2018republic,goertzel1994belief}. 
Indeed, mainstream and fringe platforms have served as \emph{de-facto} incubators of online conspiracies~\cite{adl2020}.
Notably, the impact of online conspiracy theories has been far-reaching, inciting real-world violence and influencing public health. 
The QAnon conspiracy, which gained momentum during the Trump administration, was pivotal in planning the 2021 invasion of the US Capitol \cite{riot,qanon2020}. 
At the same time, the conspiracy theories associated with COVID-19 fueled anti-vaccination sentiments and skepticism towards public health measures~\cite{puri2020social, friedrichs_fear-anger_2022}.

Mainstream platforms limit the diffusion of conspiratorial content through interventions that range from banning online communities~\cite{chandrasekharan2017you, russo2023spillover} to telling users that the information presented may be inaccurate~\cite{zannettou2021won}.
While these interventions may help curb the proliferation of conspiracy theories in online spaces \cite{russo2023understanding}, they require a fundamental technology:  ways to identify conspiratorial content accurately and at scale across various languages and cultural contexts.

In this context, we propose the \emph{Automatic Conspiracy Theory Identification (ACTI)} task.
Considering a dataset with over 25 thousand posts in Italian extracted from five Telegram channels, the ACTI consists of two subtasks: 
(i) a binary classification task where the goal is to determine if a given text piece is conspiratorial or not; and
(ii) a multi-class classification task to recognize specific conspiracy theories.

\section{Task Description}\label{sec:task_description}

The ACTI shared task comprises two subtasks, which we describe below.

\paragraph{A: Conspiratorial Content Classification.} The first subtask is determining whether a Telegram post is conspiratorial.
We consider conspiratorial texts as those that either:
(i) express the belief that influential people create major events (e.g., COVID-19) to protect their interests or 
(ii) interpret events in a way that supports the narrative of a conspiracy theory.

Note that this definition of ``conspiratorial'' is broad, as \emph{texts} may be defined as conspiratorial if they undermine commonly accepted views on societal issues.
For example, the text ``il cancro femminista sta prendendo piene'' should be classified as conspiratorial, as it subtly supports a broader theory claiming that women's rights are destroying the stability of Western societies.

\paragraph{B: Conspiracy Category Classification.} The second subtask is determining which conspiracy theory a post belongs to. 
In particular, we consider four possible conspiracy theories.

\begin{itemize}
    \item \textbf{COVID-19}: Text concerning vaccine production, 5G, and non-pharmacological interventions as a tool of control over people. Texts denying the pandemic was a real event or minimizing its importance. 

    \item \textbf{QAnon}: Texts associated with the QAnon theory.
    According to QAnon, a group of Satanic cannibalist sex abusers conspired against former U.S. President Donald Trump during his term in office. 
    This theory extended far over its original scope embodying other beliefs that support (among the others) the idea that women are enemies (hate against women) and that a powerful elite (led by public figures like Pope Francis, Queen Elizabeth, and Hillary Clinton ) is trying to organize a New World Order. 
    
    \item \textbf{Flat-Earth}: Texts associated with the claim that the earth is flat and that influential organizations hide this fact from laypeople. Usually, the flat-earth conspiracy theory is supported by pseudo-scientific evidence. 
    
    \item \textbf{Pro-Russia }: Texts associated with conspiratorial beliefs promoting Russian interests, e.g., that nazists control Ukraine's governments and army.
    
\end{itemize}

\section{Data Collection}
To gather the necessary data for the ACTI task, we employ a customized web crawler using the \textit{Selenium} and \textit{BeautifulSoup} libraries in Python. Our web crawler targets specific sources known for hosting conspiratorial content on the Telegram platform.

Specifically, we focus on a selection of Telegram channels that gained notoriety for promoting far-right ideologies and disseminating conspiracy theories. The channels we collect data from include: \emph{Qlobal-Change Italia, Basta Dittatura, Studi Scientifici Vaccini, Terra Piatta,} and \emph{Dentro La Notizia}.
For example, the channel ``Basta Dittatura'' has been actively involved in various events, including the siege of a trade union headquarters, indicating its strong affiliation with conspiratorial movements.

\begin{figure}[t]
\begin{center}

\includegraphics[width=\columnwidth]{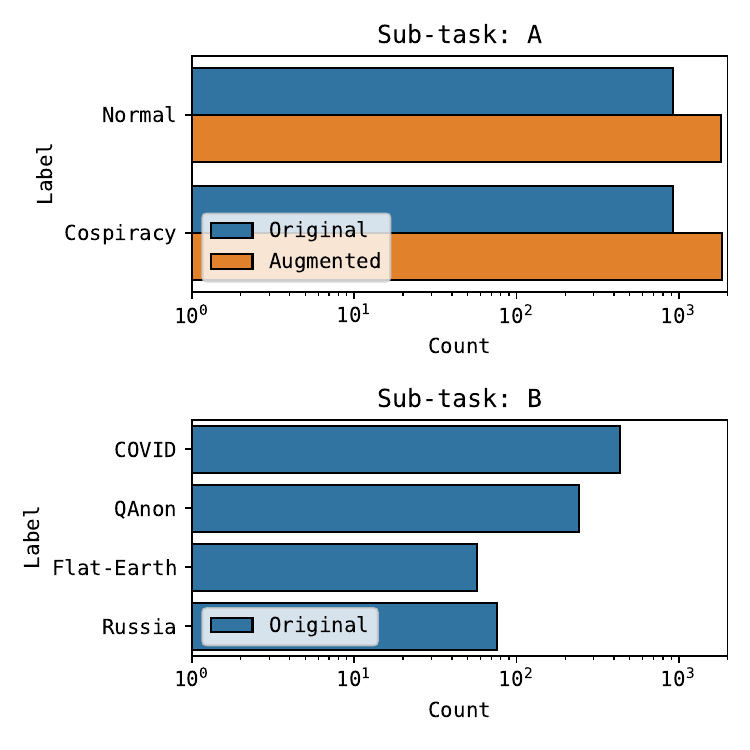}

\caption{Distribution of labels for Subtask A and Subtask B.}
\label{fig:distasks}
\end{center}
\end{figure}

Our data collection process spanned from January 1, 2020, to June 30, 2020, during which we capture and retain comments written in Italian. To ensure sufficient text for analysis, we filtered out comments with less than ten words.
We gathered a dataset comprising $25,612$ posts extracted from these five Telegram channels. We summarize statistics about our dataset in fig. \ref{fig:distasks}

\subsection{Annotation Process}

The data collection process for our study on conspiratorial content in online channels involved several steps to ensure the accuracy and relevance of the collected data.
One of the main challenges we encountered was the presence of non-conspiratorial content within the channels. 
While some comments discussed conspiratorial topics, others contained valid points or critiques regarding conspiratorial perspectives. Additionally, some comments were deemed meaningless and needed to be filtered out to maintain the integrity of the dataset.

To address this, we employed two human annotators who were responsible for labeling the comments according to three categories: ``Not Relevant,'' ``Non-Conspiratorial,'' and ``Conspiratorial.'' 
The ``Not Relevant'' label was assigned to comments that did not contribute to the discussion, while the ``Non-Conspiratorial'' label was used for comments that did not involve conspiratorial content. 
The ``Conspiratorial'' label indicated comments that contained or supported conspiratorial discussions.
For the comments labeled as ``Conspiratorial,'' we further categorized them into four subcategories: ``QAnon'', ``Covid19'', ``Russia'', and ``Flat-Earth''. These subcategories allowed us to analyze specific conspiracy theories in greater detail. The definitions of conspiratorial content are based on established studies in the field \cite{sunstein2009conspiracy, swami2011conspiracist}, ensuring consistency and clarity in our annotation process.

To assess the agreement between the annotators, we calculated inter-annotator agreement rates using Cohen's $\kappa$ coefficient. The two annotators achieved high agreement levels, with a Cohen's $\kappa$  of 0.93 for the first task and 0.86 for the second task, demonstrating the reliability of the annotation process.
To maintain data integrity, we excluded comments that did not receive the same classification from both annotators. Additionally, comments labeled ``Not Relevant'' were discarded from the dataset to focus solely on relevant conspiratorial content.

Our data collection process yielded $2{,}301$ comments for the first subtask and $1{,}110$ comments for the second subtask. This resulted in a curated dataset that provides a solid foundation for research on conspiratorial content in online discussions.

\section{Evaluation Measures}
We chose different evaluation metrics for subtasks A and B because of the distribution of the labels provided by the annotators. In particular
\paragraph{A: Conspiratorial Content Classification.} The systems submitted by participants are evaluated using the standard accuracy measures and ranked accordingly.

\paragraph{B: Conspiracy Category Classification.} Given the class imbalance for the four types of conspiracy theories we identified, we opt for using as a metric the F1-Score.
For a multi-class classification problem, we calculate the F1-score per class in a one-vs-rest manner.
We rate each class separately, computing the F1-score for each conspiracy theory in our dataset. 
To obtain a single score, we then average the per-class F1-scores. 



\paragraph{Baselines.} We follow the same methodological approach to provide a baseline for both subtasks. 
Specifically, the baselines for subtasks A and B are a Random Forest trained on a bag-of-words representation of the comments. 
In particular, we trained the random forest with $500$ estimators and validated it using a five-fold cross-validation.
These baselines achieve 0.63 accuracy for the first and  0.68 for the second subtask, respectively.

\section{Results}
 A total of fifteen teams submitted from seven institutions participated in the two tasks. Specifically, eight teams submitted for the conspiracy content classification and seven for the conspiracy category classification. 
 In total, we obtain 81 submissions. In Tables 1 and 2, we show the results for both submissions.

\begin{table}[t]
  \centering
  \label{tab:resultsa}
  \begin{tabular}{ccc}
    \hline
    \textbf{Rank} & \textbf{Team Name} & \textbf{Score} \\
    \hline
    1 & Andy P. & 0.85712 \\
    2 & extremITA & 0.85647 \\
    3 & HFI & 0.84469 \\
    4 & Flavio Giobergia & 0.83709 \\
    5 & Michael Vitali & 0.82297 \\
    6 & Giacomo Cignoni & 0.82284 \\
    7 & Selene & 0.79182 \\
    8 & Mario Graff & 0.78207 \\
    \hline
  \end{tabular}
  \caption{\textbf{Conspiratorial Content Classification:} Ranking of the eight teams joining the task. The best performing approach was obtained via Contrastive Training}
\end{table}

\subsection{Conspiratorial Content Classification}

\begin{figure}[htp]
  \centering
  \includegraphics[width=0.48\textwidth]{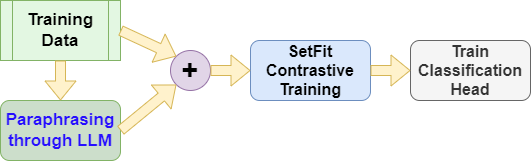}
  \caption{Overview of the methodology used by the 'Andy P.' team from the University Politehnica of Bucharest, who achieved the highest accuracy of 0.85 with five submissions}
  \label{fig:model2.png}
\end{figure}

Table 1 reports the results of the Conspiratorial Content Classification subtask, which received 40  submissions. The "Andy P." team from the University Politehnica of Bucharest achieves the highest accuracy of 0.85 with five submissions. 
Their methodology consists of an Italian language Sentence Transformer model trained it using contrastive learning. 
Due to imbalanced data, the participants integrated a data augmentation step in their classification pipeline. 
Specifically, their methodology generates synthetic data via a Large Language Model (LLM). 
These synthetic data are then used for training the model. 
Figure \ref{fig:model2.png} provides an overview of their methodology.
The second best-performing team submitted a LLM-based model as well. 
Specifically, the participants tested extremIT5 (an encoder-decoder model) and extremITLLaMA (an instruction-tuned Decoder-only Large Language Model) designed for handling Italian instructions.
While LLM-based approaches performed best, other participants developed methods based on transformers and ensembles, which achieved an accuracy of over 0.80.

\subsection{Conspiracy Category Classification}
Table 2 reports the results of the Conspiracy Category Classification task, which received 41  submissions in total. 
Once again, the ``Andy P.'' team from the University Politehnica of Bucharest achieved the highest F1-score (0.91).
Interestingly, the data augmentation process used for subtask A did increase the model performance. 
Indeed, the participants submitted the same transformer-based model trained with contrastive learning, excluding the data augmentation block. 
The second-best performing team from Tor Vergata University (Michael Vitali) achieved an F1-Score of 0.89.
They fine-tuned two BERT models, one in Italian and one multilingual, and combined them in an ensemble.
Numerous teams performed well in this task, achieving F1-scores beyond 0.80. 
Only one participant obtained a result slightly inferior to the provided baseline.

\begin{table}[t]
  \centering
  \label{tab:eresultsb}
  \begin{tabular}{ccc}
    \hline
    \textbf{Rank} & \textbf{Team Name} & \textbf{Score} \\
    \hline
    1 & Andy P. & 0.91225 \\
    2 & Michael Vitali & 0.89826 \\
    3 & HFI & 0.89476 \\
    4 & Giacomo Cignoni & 0.88534 \\
    5 & extremITA & 0.85562 \\
    6 & Flavio Giobergia & 0.83600 \\
    7 & Selene & 0.67507 \\
    \hline
  \end{tabular}
    \caption{\textbf{Conspiratorial Category Classification:} Ranking of the eight teams joining the task. The best-performing approach was obtained via Contrastive Training}

\end{table}

\section{Discussion}

A comprehensive analysis of the submitted systems reveals that most participants opted for LLMs-based models. 
Within this context, we emphasize two distinct approaches the participants employ: (i) prompting and (ii) data augmentation.
Upon thorough analysis, we find that while prompting does result in positive outcomes, the predictive capabilities of zero-shot LLMs are still inferior to systems that have been finetuned for a specific task.

\subsection{Prompting Large Language Models}
Prompting consists of providing information to a trained model to predict output labels for a task.
It is a task-agnostic approach, making it versatile and widely applicable \cite{lefebvre_rethinking_2022}. This is achieved through concise instructions, referred to as prompts, which guide the model's behavior.

The power and flexibility of prompting LLMs are well exemplified by team ExtremITA's approach: adopting a Large Language Model (LLM) to address all EVALITA tasks simultaneously.
For the ACTI task, ExtremITA is prompted with simple questions such as \emph{``Does this text talk about a conspiracy? Answer yes or no''} and \emph{``Which conspiracy theory is discussed in this text:  Covid, QAnon, Flat Earth, or Russia?''} 
This approach ranks second in subtask A of ACTI, achieving a score of 0.86 F1-score. However, it significantly drops in performance in subtask B, ranking fifth with a score of 0.85 F1-score. 

In other EVALITA tasks, ExtremITA showed significant variability in predictive capacity. 
It ranked first in eight out of twenty-five tasks, but in the remaining subtasks, it performed poorly, ranking between fifth and eleventh.
These results confirm LLMs' high potential and applicability in real-world scenarios.
However, the high variability of results shows that LLMs need help improving over models fine-tuned on specific tasks. 
Future research should focus on refining prompting techniques to improve the predictive capacity of LLMs at the single-task level.

\subsection{Augmenting Data with LLMs}

The winning team of subtasks A and B (``Andy P.'') used an approach based on data augmentation via Large Language Models and the training of sentence transformers with contrastive learning.
This approach tackles the challenge of the acquisition of conspiratorial data. 
Indeed, collecting and labeling conspiratorial data requires substantial efforts by domain specialists.  
This approach tested the possibility of leveraging LLMs to generate synthetic data and use it to train systems for automatically detecting conspiratorial content based. 
However, it is essential to note that validating the quality of data generated by LLM is an open issue within the NLP community. 
While LLMs can effectively produce synthetic content, assessing its authenticity and alignment with real-world conspiratorial beliefs is crucial.
The lower performance of the model augmented with synthetic data suggests that the quality of the generated data drastically impacts the overall model performance. 
Therefore, human evaluation is mandatory to evaluate the effectiveness of these approaches.

\section{Conclusion} 

A recent position paper \cite{basile2022evalita} asks whether EVALITA has reached its end in light of the increasing use of LLMs. 
However, based on the outcomes presented in this report, it becomes evident that the answer remains negative. The challenges posed by EVALITA tasks persist as a crucial asset in comprehending and advancing language resources and tools specifically for the Italian language. This fact is exemplified by transformer-based models' differing rankings, demonstrating the evaluation campaign's diversity and significance. 
However, the performance achieved by  LLMs is undoubtedly
pushing the limits of some tasks, especially text classification tasks.
In conclusion, while LLMs have shown great potential, EVALITA remains an essential platform for improving language tools for the Italian language.

\section*{Acknowledgments}
We thank the data annotators for their careful and valuable work. Niklas Stoehr acknowledges funding from the Swiss Data Science Center (SDSC) fellowship.

\bibliography{bibliography}

\end{document}